\documentclass[letterpaper]{article} 
\usepackage[preprint]{aaai2027}  
\usepackage[hyphens]{url}  
\usepackage{graphicx} 
\usepackage{natbib}  
\usepackage{caption} 
\usepackage{algorithm}
\usepackage{algorithmic}

\usepackage{newfloat}
\usepackage{listings}
\DeclareCaptionStyle{ruled}{labelfont=normalfont,labelsep=colon,strut=off} 
\floatstyle{ruled}
\newfloat{listing}{tb}{lst}{}
\floatname{listing}{Listing}

\usepackage{booktabs}

\usepackage{amsmath}
\usepackage{tabularx}
\usepackage{multirow}
\usepackage[most]{tcolorbox}    

\newtcolorbox{promptbox}[1]{
    fontupper=\small,
    colback=white,    
    colframe=black,  
    fonttitle=\bfseries,     
    title={#1},              
    boxrule=0.5pt,           
    left=5pt, right=5pt, top=5pt, bottom=5pt
}

\title{Select-And-Extract: A Lightweight Plugin for Retrieval-Augmented Generation}
\author {
    Chenming Tang\textsuperscript{\rm 1},
    Jiawei Han\textsuperscript{\rm 2}
}

\affiliations {
    \textsuperscript{\rm 1}Peking University,
    \textsuperscript{\rm 2}University of Illinois Urbana-Champaign\\
    tangchenming@stu.pku.edu.cn,
    hanj@illinois.edu
}

\begin{document}

\maketitle

\begin{abstract}
Retrieval-augmented generation (RAG) for language model (LM) systems fundamentally has two failure modes: retrieval failure and reading failure. The former fails to recall the right pieces of information from the external corpus, and the latter fails to produce the correct answer although the right information is retrieved. Some methods perform structured indexing for retrieval failure, but may suffer from limited generalization of the fixed structures. Some methods perform query-time structuring for reading failure, but typically require a lot of LM calls and rely heavily on the LM's capability. To this end, we propose Select-ANd-Extract (SANE), a simple yet effective plugin for RAG. For the retrieval failure, we retrieve a wide set of candidates with a semantic retriever, and leverage the LM to select the top candidates based on their synopses, which yields better recall than the original retriever. For the reading failure, we perform blueprint-guided query-time evidence extraction, which allows the generator LM to use only compact and structured key information so that it can perform better reasoning. Empirical results confirm that SANE brings solid improvements, while only introducing modest extra overhead. As a lightweight plugin for RAG, SANE offers a simple alternative to heavier approaches, and suggests a high-performance RAG framework need not be overly complex.
\end{abstract}

\begin{links}
    \link{Code}{https://github.com/JamyDon/SANE}
\end{links}

\section{Introduction}

Language models (LMs) encode all the knowledge in their parameter space. Retrieval-augmented generation (RAG) enhances LM systems by retrieving knowledge from an external corpus~\cite{RAG,rag-survey-24}. Fundamentally, RAG has two basic failure modes: retrieval failure and reading failure. The former means the retriever fails to recall the right information, typically due to the limitation of the retrieval approach. The latter means the generator LM fails to produce the correct answer even when the right information is retrieved, usually due to the limited capability of the LM, or the coarse and noisy representation of the information.

Existing RAG approaches address these failures from different stages of the pipeline. For retrieval, prior research improves the retriever~\cite{DPR,ColBERT,Hybrid-RAG}, rewrites the query~\cite{Query-Rewriting}, expands the document~\cite{Doc2Query}, and builds topology indexes~\cite{RAPTOR,GraphRAG}. However, with the development of embedding models~\cite{qwen-embedding}, semantic embeddings capture increasingly rich information, reducing the marginal benefit of some retrieval-time enhancements. For reading, the mainstream is to structure the information, either at index time through corpus-level ingestion~\cite{RAPTOR,HippoRAG,HippoRAG2}, or at query time through query-specific structuring~\cite{StructRAG,RAS,SARG,SLIDERS}. Due to the powerful capability of LMs, most of the existing methods employ LMs to perform high-quality operations, requiring a lot of LM calls for each query, which is unfriendly to cost-constrained application scenarios. Meanwhile, they rely heavily on the capability of LMs and may not perform well without specific training or advanced large-scale LMs.

To this end, we introduce \textbf{S}elect-\textbf{AN}d-\textbf{E}xtract (SANE), a simple, lightweight, and training-free plugin for RAG that targets its two fundamental failure modes while overcoming the limitations of existing methods. For the retrieval failure, we introduce synopsis-based candidate selection to help the recall, which leverages the LM to select top candidates based on the synopses including easily obtained features of a wide set of candidates retrieved by a semantic retriever. For the reading failure, we propose blueprint-guided evidence extraction for better information representation at read time, which calls the LM once with the raw content of all the selected candidates, the query, and a few canonical schema blueprints as guidance, to perform query-specific evidence extraction. Finally, the generator LM only sees the query and the compact structured information to perform the reasoning. The contribution of SANE is not selection or extraction alone, but a low-cost query-time augmentation mechanism that combines synopsis-based candidate selection with blueprint-guided evidence extraction.

\begin{figure*}[htbp]
\centering
\includegraphics[width=0.95\textwidth]{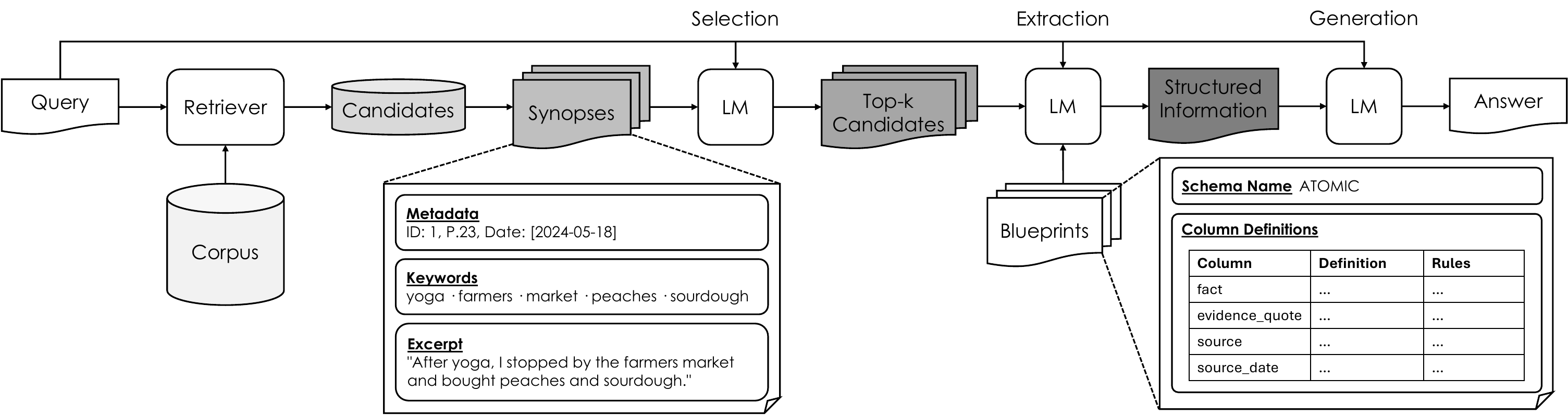}
\caption{Overview of the SANE pipeline. Just three LM calls: selection, extraction, and generation.}
\label{fig:method}
\end{figure*}

We evaluate SANE against other RAG enhancement methods across four benchmarks, covering multi-hop QA and long-document QA tasks. The empirical results show that SANE consistently brings solid improvements with modest overhead. Quantitatively, across all the benchmarks and two backbone LMs, SANE outperforms the strongest baseline by \textbf{+9.97} points on average. Ablation studies validate the effectiveness of all the components of our design, while additional analysis further confirms the advantage of SANE. Overall, our empirical evaluation demonstrates that SANE offers a simple alternative to heavier or more complex approaches.

Our contributions are threefold:
\begin{itemize}
    \item We introduce SANE, a lightweight yet effective plugin for RAG that improves the retrieval recall and read-time information representation.
    \item We empirically validate that SANE brings solid improvements for RAG with modest extra overhead, offering a lightweight alternative to heavier approaches.
    \item We show that the simple two-step plugin is already effective, suggesting that an advanced RAG framework need not be overly complex.
\end{itemize}


\section{Method}\label{sec:method}

The overview of SANE is shown in Figure~\ref{fig:method}, and a formal algorithm description of SANE is provided by Algorithm~\ref{alg:SANE}.

\subsection{Problem Formulation}
Given a RAG-enhanced LM system $\mathcal{S}$ driven by a generator LM $\theta$, the system generates the response $a$ based on the query $q$ and the external corpus $D$:
\begin{equation}
    a=\mathcal{S}(q,D)=\theta(q,d),
\end{equation}
where $d$ is the information retrieved by the retrieval module $\mathcal{R}$ from $D$ for $q$. Traditional embedding-based RAG stores $D$ as raw content (\textit{e.g.}, chunked documents) and implements $\mathcal{R}$ as retrieving the most similar $k$ pieces for $q$:
\begin{equation}
    d=\mathcal{R}(q,D,k)=\mathop{\text{Top-}k}_{c\in D}(\mathrm{sim}(q,c)).
\end{equation}

\subsection{Synopsis-Based Candidate Selection}
Although embedding models have become powerful, they can still suffer from false positive distractors, which are semantically similar but not really relevant. Therefore, SANE leverages the LM to better select the recalled candidates, which can be more flexible compared to the similarity score.

Concretely, we retrieve a large set $\hat{C}$ of $w$ candidates ($w>k$) using the similarity-based retriever to ensure a high upper bound for recall while introducing no extra overhead:
\begin{equation}
\hat{C}=\mathcal{R}(q,D,w) =\{c_i\}_{i=1}^w.
\end{equation}

The raw content of these candidates can be very long. Therefore, we leverage a few cheap features like pre-existing metadata $\mu$ (\textit{e.g.}, ID, date, page number), TF-IDF keywords $\kappa$, and query-aware lexical-overlap-based excerpts $\epsilon$ (\textit{i.e.}, key sentence) to generate a short synopsis $z$ for each candidate $c$, which is much shorter than the raw content:
\begin{equation}
    \mu_i = \mathrm{Metadata}(c_i),
\end{equation}
\begin{equation}
    \kappa_i = \mathrm{Keywords}(c_i, \hat{C}),
\end{equation}
\begin{equation}
    \epsilon_i = \mathrm{Excerpt}(c_i, q),
\end{equation}
\begin{equation}
    z_i = (\mu_i, \kappa_i, \epsilon_i).
\end{equation}
These features are all rule-based and fast to compute, free of any LMs or trained policy. For the keywords, we filter common stopwords and compute IDF over all the $w$ candidates.

Finally, we call the LM $\theta$ to select the top-$k$ candidates that are most likely to be useful for $q$:
\begin{equation}
    C=\theta(q, \{z_i\}_{i=1}^w,k).
\end{equation}
The LM call is not very costly since the synopses are short.

\subsection{Blueprint-Guided Evidence Extraction}

\begin{table*}[htbp]
\small
  \centering
    \begin{tabularx}{\textwidth}{>{\hsize=.25\hsize\linewidth=\hsize}X
  >{\hsize=.52\hsize\linewidth=\hsize}X>{\hsize=.6\hsize\linewidth=\hsize}X>{\hsize=.5\hsize\linewidth=\hsize}X}
    \toprule
    \textbf{Schema} & \textbf{Information Structure} & \textbf{Use When} & \textbf{Columns} \\
    \midrule
    ATOMIC & Independent facts — no row relates to any other & What / Who / Where / Define / Find / Yes-no questions & fact, evidence quote, source, source date \\
    \midrule
    COMPARATIVE & Aligned dimensions — entities compared on shared attributes & Compare / Difference / Versus / Which is larger-better-faster / Rank & entity, dimension, value, unit, evidence quote, source \\
    \midrule
    TEMPORAL & Sequenced events — rows ordered by occurrence time & When / First / Last / {Most recent} / Timeline / Before-After / Sequence & event, position, resolved date, date precision, evidence quote, source \\
    \bottomrule
    \end{tabularx}%
  \caption{Blueprints serving as in-context structural guides for the evidence extraction.}
  \label{tab:blueprints}%
\end{table*}%

Traditional RAG presents the raw content of the recalled chunks to the generator LM, which can be long yet noisy. Meanwhile, many query-time structuring methods require multiple extra LM calls for each query, which can be extremely expensive and rely heavily on the LM's capability. Therefore, we propose to extract the key information for the query by structuring the recalled chunks with only one extra LM call. This presents the most informative things in a structured way while not requiring multiple LM calls.

Instead of extracting evidence based on fixed schemas, we let the LM perform extraction using customized schemas based on the query, which ensures flexibility and adaptivity of the structuring. We include a set of blueprints $B$ in the prompt as guidance, each including a general schema as a demonstration. These pre-defined schemas, as shown in Table~\ref{tab:blueprints}, can be generally applied in common cases, covering frequent usage. They are mainly used as in-context demonstrations that guide the LM to design and extract the schema, instead of fixed rules to constrain the LM's extraction. In this way, the LM benefits from certain guidance while having the freedom to use adaptive schemas based on the query.

We only call the LM once to extract the structured evidence as the final information presented to the generator LM from all the retrieved chunks, guided by the blueprints:
\begin{equation}
    d=\theta(q, C, B).
\end{equation}
Finally, we pass the structured evidence $d$ to the generator LM and perform the final generation for the answer $a$:
\begin{equation}
    a=\theta(q,d).
\end{equation}

\begin{algorithm}[htbp]
\caption{SANE: Select-And-Extract}
\label{alg:SANE}
\textbf{Input}: query $q$, corpus $D$\\
\textbf{Parameter}: wide pool size $w$, final budget $k$, blueprint set $B$, retriever $\mathcal{R}$, backbone LM $\theta$\\
\textbf{Output}: answer $a$
\begin{algorithmic}[1]
\STATE Retrieve a wide candidate pool $\hat{C} \leftarrow \mathcal{R}(q, D, w)$.
\FOR{each candidate $c_i \in \hat{C}$}
    \STATE Obtain metadata if applicable $\mu_i \leftarrow \mathrm{Metadata}(c_i)$.
    \STATE Compute TF-IDF keywords $\kappa_i \leftarrow \mathrm{Keywords}(c_i, \hat{C})$.
    \STATE Select a query-aware excerpt $\epsilon_i\leftarrow \mathrm{Excerpt}(c_i, q)$.
    \STATE Construct synopsis $z_i \leftarrow (\mu_i, \kappa_i, \epsilon_i)$.
\ENDFOR
\STATE Select candidates $C \leftarrow \theta(q, \{z_i\}_{i=1}^{w}, k)$.
\STATE Extract structured evidence $d \leftarrow \theta(q, C, B)$.
\STATE Generate answer $a \leftarrow \theta(q, d)$.
\STATE \textbf{return} $a$
\end{algorithmic}
\end{algorithm}

\section{Experimental Setup}

\subsection{Benchmarks}

\begin{table}[htbp]
  \centering
    \begin{tabular}{llrr}
    \toprule
    \textbf{Benchmark} & \textbf{Task} & \multicolumn{1}{l}{\textbf{\#Inst.}} & \textbf{\#Tok/Inst.} \\
    \midrule
    HotpotQA & Multi-hop & 7,405 & 1K \\
    MuSiQue & Multi-hop & 2,417 & 2K \\
    FinanceBench & Long-document & 150   & 115K \\
    LongMemEval & Long-context & 500   & 1018K \\
    \bottomrule
    \end{tabular}%
  \caption{Benchmark statistics. ``\#Inst.'' and ``\#Tok/Inst.'' refer to the instance count and tokens per instance, respectively.}
  \label{tab:data-stat}%
\end{table}%

We evaluate on two different tasks. For multi-hop QA, we use HotpotQA~\cite{HotpotQA} and MuSiQue~\cite{MuSiQue}, which require gathering information from multiple sources and reasoning on it. For long-document/long-context QA, we use FinanceBench~\cite{FinanceBench} with long financial filings, and the M subset of LongMemEval~\cite{LongMemEval} with long conversation histories (over one million tokens). Table~\ref{tab:data-stat} lists the benchmark statistics.

\subsection{Evaluation Metrics}
For end-to-end performance, we follow the official metrics of the benchmarks. For HotpotQA and MuSiQue, we report exact match (EM) as the primary metric, paired with F1 score (F1). For FinanceBench and LongMemEval, we report LLM-judged accuracy by DeepSeek-V4-Flash~\cite{DeepSeek-V4}. We also report efficiency metrics: total LM token consumption (in millions), number of LM calls per instance, and per-instance latency (in seconds).

\subsection{Baselines}
We evaluate the following methods as baselines: (1) Embedding: vanilla RAG using embedding-based semantic search, which retrieves the top-$k$ similar candidates and passes the retrieved content to the generator LM; (2) RAPTOR~\cite{RAPTOR}: recursively clusters and summarizes the documents and builds hierarchical indexes; (3) HippoRAG 2~\cite{HippoRAG2}: builds neurobiologically inspired knowledge graphs, and retrieves through graph traversal and Personalized PageRank propagation; (4) StructRAG~\cite{StructRAG}: selects a structure type (graphs, tables, \textit{etc.}) and extracts the retrieved content into the chosen structure; (5) RAS \cite{RAS}: iteratively retrieves evidence and constructs a query-specific knowledge graph for answering.

\subsection{Backbone Models}
We evaluate multiple backbone LMs. For large-scale LMs, we employ DeepSeek-V4-Flash~\cite{DeepSeek-V4}. For small-scale LMs, we employ Qwen3.5-9B~\cite{Qwen3}. For embedding, we employ Qwen3-Embedding-0.6B \cite{qwen-embedding}, which is a moderate-scale embedding model.

\begin{table*}[htbp]
  \centering
    \begin{tabular}{lccccccc}
    \toprule
    \multirow{2}[2]{*}{\textbf{Method}} & \multicolumn{2}{c}{\textbf{HotpotQA}} & \multicolumn{2}{c}{\textbf{MuSiQue}} & \multirow{2}[2]{*}{\textbf{FinanceBench}} & \multirow{2}[2]{*}{\textbf{LongMemEval}} & \multirow{2}[2]{*}{\textbf{Avg.}} \\
    \cmidrule(lr){2-3}\cmidrule(lr){4-5} 
          & \textbf{EM} & \textbf{F1} & \textbf{EM} & \textbf{F1} &       &  \\
    \midrule
    \multicolumn{8}{c}{\textit{DeepSeek-V4-Flash}} \\
    \midrule
    Embedding & 50.18 & 64.44 & 28.09 & 41.03 & 58.67 & 43.2 & 45.04\\
    RAPTOR & 47.08 & 60.75 & 25.07 & 37.43 & 45.33     & 29.8 & 36.82 \\
    HippoRAG 2 & 54.44 & 69.35 & 32.77 & 46.54 & 58.67 & 49.6 & 48.87\\
    StructRAG & 45.33 & 58.80  & 21.43 & 33.60  & 60.67 & 38.6 & 41.51\\
    RAS$_\text{lite}$ & 40.34 & 53.95 & 26.73 & 39.59 & 32.67 & 15.2 & 28.74\\
    SANE (Ours) & \textbf{59.96} & \textbf{75.81} & \textbf{40.26} & \textbf{53.78} & \textbf{75.33} & \textbf{57.4} & \textbf{58.24}\\
    \midrule
    \midrule
    \multicolumn{8}{c}{\textit{Qwen3.5-9B}} \\
    \midrule
    Embedding & 48.40 & 60.55 & 26.52 & 37.65 & 57.33 & 40.0 & 43.06\\
    RAPTOR & 43.24 & 55.26 & 23.67 & 34.73 & 34.67     & 25.4 & 31.75 \\
    HippoRAG 2 & 54.23 & 66.95 & 35.17 & 46.40 & 56.67 & 43.2 & 47.32 \\
    StructRAG & 36.60 & 48.29 & 12.74 & 20.73 & 43.33 & 18.8 & 27.87\\
    RAS$_\text{lite}$ & 21.49 & 30.14 & \phantom{0}5.38  & 14.35 & 24.00 & \phantom{0}9.2 & 15.02\\
    SANE (Ours) & \textbf{61.67} & \textbf{76.45} & \textbf{44.85} & \textbf{57.49} & \textbf{71.33} & \textbf{56.4} & \textbf{58.56}\\
    \bottomrule
    \end{tabular}%
  \caption{The main evaluation results. The highest scores are bolded. ``Avg.'' computes the averaged primary scores (exact match for HotpotQA and MuSiQue, and accuracy for FinanceBench and LongMemEval) across all the four benchmarks.}
  \label{tab:main}%
\end{table*}%

\subsection{Other Implementation Details}

The retrieval budgets $w$ and $k$ are 10 and 3 for HotpotQA, and 20 and 5 for other benchmarks. The chunking is based on natural boundaries for HotpotQA (paragraph), MuSiQue (paragraph) and LongMemEval (history session), and uses a chunk size of 2,048 for FinanceBench. For a higher throughput, we run 10 and 20 concurrent instances for the DeepSeek and Qwen backbones, respectively. To minimize LM-side variance, we set the temperature to 0 for all the LM calls.

For SANE, if the selection phase selects fewer than $k$ candidates, we pad the remaining slots with unselected top embedding candidates to ensure a consistent retrieval budget. For StructRAG, we train Qwen3.5-9B as the router for both backbone LMs. For RAS, we evaluate the training-free version (named as RAS$_\text{lite}$) by directly prompting the LMs, because the training of RAS is highly resource-consuming.

More technical details, like the detailed LM prompts, can be found in the supplementary material.

\begin{table*}[htbp]
\setlength{\tabcolsep}{2pt}
  \centering
    \begin{tabular}{lllccccccccc}
    \toprule
    \multirow{2}[2]{*}{\textbf{Method}} & \multirow{2}[2]{*}{\textbf{LM Input}} & \multirow{2}[2]{*}{\textbf{LM Call}} & \multicolumn{3}{c}{\textbf{MuSiQue}} & \multicolumn{3}{c}{\textbf{FinanceBench}} & \multicolumn{3}{c}{\textbf{LongMemEval}} \\
\cmidrule(lr){4-6}\cmidrule(lr){7-9}\cmidrule(lr){10-12}          &       &       & \textbf{Token} & \textbf{ Call } & \textbf{Latency} & \textbf{Token} & \textbf{ Call } & \textbf{Latency} & \textbf{Token} & \textbf{ Call } & \textbf{Latency} \\
    \midrule
    Embedding & $k\cdot L$    & $1$     & \phantom{0}2.1   &      \phantom{0}1  & \phantom{0}0.8   & \phantom{0}0.5   &        \phantom{00}1  & \phantom{0}15.4   & \phantom{000}4.9   &      \phantom{00}1  & \phantom{0}27.8\phantom{*} \\
    RAPTOR & $\frac{b}{b-1}N\cdot L$   & $\frac{1}{b-1}N+1$ & \phantom{0}9.1   &    \phantom{0}6  & 10.1  & 19.0     &  \phantom{0}30    & 101.4     & \phantom{00}56.2     &  \phantom{0}31    & \phantom{0}95.0\phantom{*} \\
    HippoRAG 2 & $(2N+k)\cdot L$ & $2N+2$  & 81.2  &  42  & 40.6  & 96.9  &   594  & 618.6  & 1432.4     &  947    & 238.8* \\
    StructRAG & $k\cdot L+sk\cdot l_\text{struct}$ & $k+s+3$ & 13.7  &    \phantom{0}8  & 15.3  & \phantom{0}2.4   &     \phantom{0}10  & \phantom{0}65.4   & \phantom{000}9.0   &   \phantom{00}9  & \phantom{0}45.4\phantom{*} \\
    RAS$_\text{lite}$ & $tk\cdot L+t^2\cdot l_\text{struct}$ & $2t+2$  & \phantom{0}4.6   &      \phantom{0}2  & \phantom{0}2.1   & \phantom{0}0.2   &        \phantom{00}2  & \phantom{00}0.9   & \phantom{000}0.7   &   \phantom{00}2  & \phantom{00}0.9\phantom{*} \\
    SANE (Ours) & $w\cdot l_\text{synop}+k\cdot L+l_\text{struct}$ & $3$     & 10.7  &      \phantom{0}3  & \phantom{0}5.2  & \phantom{0}1.3   &        \phantom{00}3  & \phantom{0}18.3   & \phantom{000}7.8   &   \phantom{00}3  & \phantom{0}32.6\phantom{*} \\
    \bottomrule
    \end{tabular}%
  \caption{Cost analysis with Qwen3.5-9B as the LM, focusing on total token consumption (million), per-instance LM calls, and per-instance latency (second). ``LM Input'' and ``LM Call'' are formal estimations of the LM input tokens and number of LM calls per instance. $L$ is the typical chunk size, $N$ is the number of chunks, $k$ and $w$ are retrieval budgets defined in ``Method'', $b$ is the branching factor of RAPTOR, $s$ is the number of sub-queries of StructRAG, $t$ is the number of iterations of RAS, $l_\text{struct}$ is the typical size of structured information, and $l_\text{synop}$ is the typical size of each synopsis of SANE. ``*'' indicates that HippoRAG 2 uses intra-instance concurrency of 100 for primary information extraction, instead of the inter-instance concurrency of 20 used in other settings, due to contention issues on LongMemEval.}
  \label{tab:cost}%
\end{table*}%

\section{Results and Analysis}
\subsection{Main Results}

The main results are presented in Table~\ref{tab:main}. Overall, our proposed SANE secures the highest scores across all the benchmarks with both backbone LMs, with averaged gains of \textbf{+6.48}, \textbf{+8.59}, \textbf{+14.33}, and \textbf{+10.50} over the second strongest method on the four benchmarks, respectively.

Other compared methods exhibit certain limitations. Embedding, with the help of a modern embedding model, achieves relatively balanced performance, but is constantly outperformed by SANE. Index-time structuring methods like RAPTOR sometimes even underperform Embedding because the structures are fixed and may not be as general as the embedding, especially when today's embedding model is getting more powerful. Query-time structuring methods like StructRAG and RAS rely heavily on specially trained models or the LM's capability, and perform poorly especially when the LM is smaller and less competitive. Overall, HippoRAG 2 is the strongest baseline, with solid improvements over the balanced Embedding in many cases, but underperforms SANE even when costing significantly more tokens and time, which will be specified in the next section.

\subsection{Cost Analysis}

We perform cost analysis both formally and empirically in Table~\ref{tab:cost}. SANE introduces modest extra overhead over Embedding, with only two extra LM calls and not too many tokens for the selection and extraction. Its cost is dominated by the retrieval budgets $w$ and $k$ rather than the number of chunks $N$, since SANE needs no extra index-time operations. Its overhead is more marginal on longer benchmarks like FinanceBench and LongMemEval, where $N$ is very large.

Among index-time structuring methods, RAPTOR is efficient on shorter data like MuSiQue where the dominating $N$ is small, with the overall cost on par with SANE, but its benchmark performance is much worse and it costs much more on longer benchmarks. HippoRAG 2, due to the dominating chunk count $N$, is the heaviest method, especially on FinanceBench and LongMemEval, where the LM cost is approximately \textbf{two orders of magnitude greater} than SANE.

Query-time structuring methods cost less on long benchmarks since they do not perform complex indexing and are thus not dominated by the large $N$. StructRAG has a modestly higher overall cost than SANE due to more LM calls for sub-queries. Meanwhile, RAS, with poor performance when training-free, has a low cost because the LM decides no retrieval required in many cases due to incorrect judgment, which causes the LM to directly answer the query without extra operations (\textit{i.e.}, the iteration count $t$ is zero).

The cost analysis confirms that SANE balances both performance and efficiency, requiring modest overhead while performing best. It also scales well to longer data.

\subsection{Cost-Effectiveness of Heavy Methods}

\begin{table}[htbp]
\setlength{\tabcolsep}{1.5pt}
  \centering
    \begin{tabular}{lrrrr}
    \toprule
    \textbf{Method} & \textbf{Score} & \textbf{Token (M)} & \textbf{LM Call} & \textbf{Latency (s)} \\
    \midrule
    Embedding & 57.33 & 0.5   &          1  & 15.4\phantom{*} \\
    GraphRAG$_\text{fast}$ & 40.00 & 267.9 & 268 & 2,197.3\phantom{*} \\
    GraphRAG$_\text{standard}$ & 50.00 & 684.0 &   2,029  & 5,355.9\phantom{*} \\
    SLIDERS & 48.00 & 111.5 &     260  & 399.9* \\
    SANE (Ours) & \textbf{71.33} & 1.3   &          3  & 18.3\phantom{*} \\
    \bottomrule
    \end{tabular}%
  \caption{Extended comparison results on FinanceBench. ``*'' means SLIDERS uses a different concurrency setting (5 intra-instance and 4 inter-instance; other methods solely use 20 inter-instance) due to bounded resources.}
  \label{tab:extended}%
\end{table}%

Some methods are extremely heavy, so we evaluate them only on FinanceBench within our budget, with Qwen3.5-9B as the backbone LM. These methods include: (1) GraphRAG~\cite{GraphRAG}: builds an entity-relation graph and generates community summaries; the ``fast'' and ``standard'' variants extract information with spaCy models and LMs, respectively; (2) SLIDERS~\cite{SLIDERS}: a query-time structuring method that extracts information into databases, and applies a SQL coding agent to reason and generate the answer.

Table~\ref{tab:extended} presents the results. SANE secures the highest accuracy, with significantly less overhead (typically two orders of magnitude less). All the heavy methods require many extra LM calls and rely heavily on the LM's capability. Therefore, they consume significantly more tokens and time, while even underperforming Embedding when the backbone LM is a relatively small 9B one. This again confirms SANE is a simple yet effective method, less sensitive to backbone LM capability and requiring much less cost.

\subsection{Ablation Study}

\begin{table}[htbp]
\setlength{\tabcolsep}{4.5pt}
  \centering
    \begin{tabular}{lrrrrr}
    \toprule
    \textbf{Setting} & \textbf{MQ} & \textbf{FB} & \textbf{LME} & \textbf{Avg.} & \textbf{$\Delta$} \\
    \midrule
    SANE (Default) & \textbf{44.85} & \textbf{71.33} & \textbf{56.40} & \textbf{57.53} & - \\
    \midrule
    \multicolumn{6}{l}{\textit{General Ablation on Components}} \\
    \;w/o selection & 31.32 & 58.00 & 44.00 & 44.44 & -13.09 \\
    \;w/o extraction & 38.44 & 68.00 & 52.40 & 52.95 & -4.58 \\
    \midrule
    \multicolumn{6}{l}{\textit{Focused Ablation on Synopsis}} \\
    \;w/o synopsis & \textbf{44.85} & 70.00 & 54.40 & 56.42 & -1.11 \\
    \;w/o metadata & 44.02 & 70.67 & 55.00 & 56.56 & -0.96 \\
    \;w/o excerpt & 39.22 & 70.00 & 52.20 & 53.81 & -3.72 \\
    \;w/o keyword & 44.31 & 70.00 & 52.80 & 55.70 & -1.82 \\
    \midrule
    \multicolumn{6}{l}{\textit{Focused Ablation on Blueprint}} \\
    \;w/o blueprint & 40.17 & \textbf{71.33} & 52.20 & 54.57 & -2.96 \\
    \bottomrule
    \end{tabular}%
  \caption{Ablation results on MuSiQue (MQ), FinanceBench (FB), and LongMemEval (LME), with Qwen3.5-9B as the backbone LM. ``$\Delta$'' shows the decrease of the averaged score.}
  \label{tab:ablation}%
\end{table}%

We perform ablation studies on SANE (Table~\ref{tab:ablation}).

The synopsis-based selection is especially important, with a significant decrease of \textbf{13.09} averaged points across the compared benchmarks when removed. The ``w/o synopsis'' setting directly prompts the LM with the full raw content of the candidates and serves as a proxy for zero-shot LM-based reranking. Compared with the synopsis, the raw content can be long and noisy and may confuse the LM especially when the chunk is large on long-document benchmarks, as shown by the degraded performance. It is also noteworthy that this full-content setting consumes significantly more context tokens for the selection phase (\textbf{2$\times$} and \textbf{14$\times$} tokens on FinanceBench and LongMemEval, respectively). In addition, the three parts of the synopsis (metadata, excerpt, and keyword) all contribute, and the excerpt contributes the most with \textbf{3.72} averaged points.

Meanwhile, the blueprint-guided extraction is also indispensable, with a degradation of \textbf{4.58} averaged points when removed. We also ablate the blueprints by extracting the information with a trivial prompt without any specific blueprint or schema guidance, and show that there is a drop of \textbf{2.96} averaged points when the blueprint guidance is disabled.

The above ablation results confirm that each part of our proposed SANE is effective and indispensable.

We also evaluate the effect of retrieval hyperparameters, with the results in the technical supplement. The results suggest that the current $w$ and $k$ values are near-optimal, and SANE brings consistent gains over vanilla RAG under all the evaluated hyperparameter values.

\subsection{Recall Analysis}

\begin{table}[htbp]
\setlength{\tabcolsep}{4pt}
  \centering
    \begin{tabular}{lcrrrr}
    \toprule
    \textbf{Recall} & \textbf{\#} & \textbf{MQ} & \textbf{FB} & \textbf{LME} & \textbf{Avg.} \\
    \midrule
    Embedding Top-20 & $w=20$     & 100.0 & 60.8  & 68.8  & 76.5 \\
    \midrule
    Embedding Top-5 & $k=5$     & 69.4  & 38.3  & 48.6  & 52.1 \\
    SANE Selected & $k=5$     & \textbf{88.7} & \textbf{51.1} & \textbf{67.3} & \textbf{69.0} \\
    \bottomrule
    \end{tabular}%
  \caption{Recall analysis with Qwen3.5-9B on MuSiQue (MQ), FinanceBench
(FB), and LongMemEval (LME).}
  \label{tab:recall}%
\end{table}%

\begin{figure*}[htbp]
\centering
\includegraphics[width=1.0\textwidth]{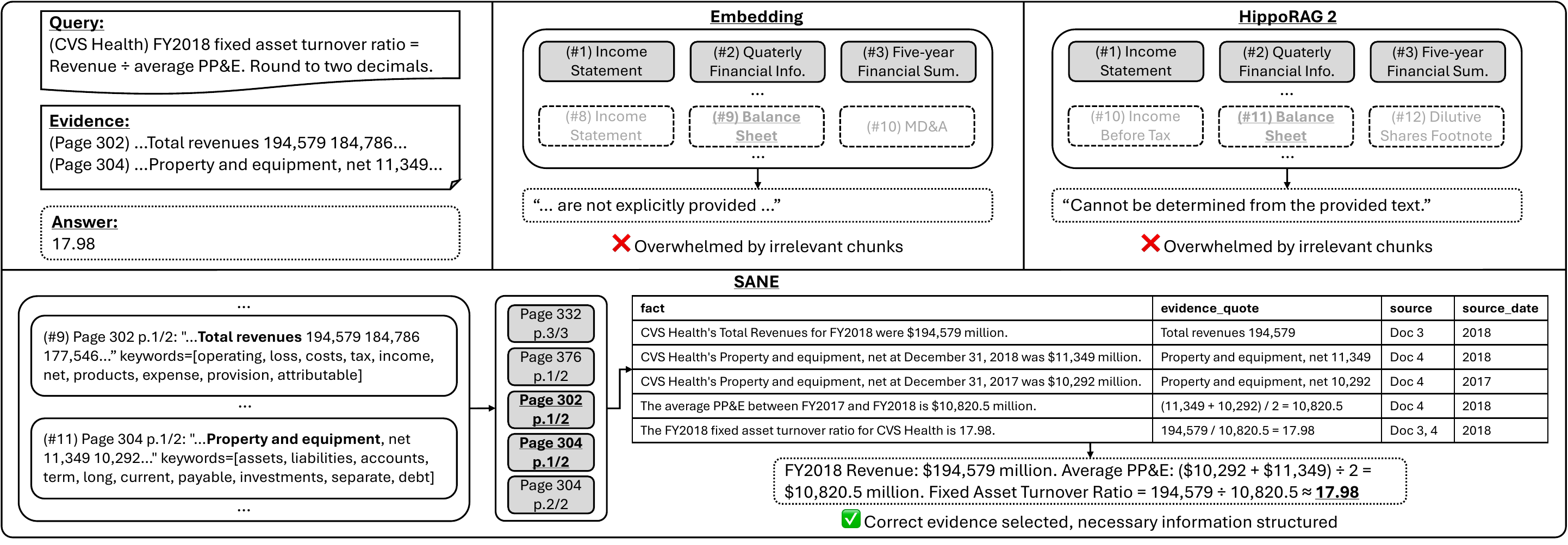}
\caption{A concrete case from FinanceBench comparing embedding, HippoRAG 2, and SANE.}
\label{fig:case}
\end{figure*}

We perform recall analysis to confirm that the synopsis-based selection phase adds value. Table~\ref{tab:recall} compares the recall@$k$ of the native top-$k$ of the embedding and the selected $k$ candidates on benchmarks with ground truth evidence annotations against the wide recall@$w$ of the embedding. SANE consistently brings significantly higher recall@$k$ across all the three benchmarks (\textbf{+19.3}, \textbf{+12.8}, and \textbf{+18.7}, respectively), bridging the gap between embedding's recall@$k$ and recall@$w$. This confirms the synopsis-based selection makes a positive contribution by improving the recall.

\subsection{Extraction Quality Audit}

We quantitatively evaluate the quality of the blueprint-guided evidence extraction on FinanceBench with Qwen3.5-9B as the backbone. We prompt an advanced LM, DeepSeek-V4-Pro \cite{DeepSeek-V4}, to annotate each row in the extracted information. The results suggest that most extracted rows (81.1\%) are fully faithful, while there is also a non-trivial ratio of unsupported rows (13.9\%) suffering from hallucination. We acknowledge this is a limitation of the single-pass extraction, and leave the exploration of more faithful single-pass extraction to future work. The full breakdown is reported in the technical supplement.

\subsection{Robustness to Backbone LMs}

\begin{table}[htbp]
\setlength{\tabcolsep}{4pt}
  \centering
    \begin{tabular}{lrrr}
    \toprule
    \textbf{LM} & \textbf{Embedding} & \textbf{SANE} & \textbf{$\Delta$} \\
    \midrule
    Qwen3.5-4B & 40.36 & 53.92 & +13.56 \\
    Qwen3.6-35B-A3B-FP8 & 46.92 & 60.89 & +13.97 \\
    DeepSeek-V4-Pro & 48.24 & 60.56 & +12.32 \\
    \bottomrule
    \end{tabular}%
  \caption{Averaged results with different LMs as the backbone.}
  \label{tab:models}%
\end{table}%

We evaluate both Embedding and SANE on all the four benchmarks with other backbone LMs to confirm the advantage of SANE generalizes across different LMs. Table~\ref{tab:models} presents the averaged scores across all the four benchmarks for both methods. SANE consistently brings substantial improvement over the vanilla RAG with all the backbone LMs. This suggests that SANE benefits RAG whenever the LM is small (4 billion parameters for Qwen3.5-4B) or large (1.6 trillion total parameters for DeepSeek-V4-Pro) and the architecture is dense (Qwen3.5 series) or sparse (mixture-of-experts for Qwen3.6-35B-A3B-FP8 and the DeepSeek series).

\subsection{Case Study}

To demonstrate the advantage of SANE concretely, we provide a case from FinanceBench comparing Embedding, HippoRAG 2, and SANE, with Qwen3.5-9B as the backbone LM (Figure~\ref{fig:case}). The query asks about CVS Health's fixed asset turnover ratio of fiscal year 2018, which equals revenue (on page 302) divided by average PP\&E (on page 304).

The embedding-based retrieval is overwhelmed by similar but irrelevant chunks, while HippoRAG 2's Personalized PageRank surfaces income statement and summary pages. Therefore, the ground truth evidence chunks fall out of the top-$k$ (highest ranked 9 and 11, respectively). As a result, the generator LM cannot answer due to the missing information.

On the contrary, SANE answers correctly. In the selection phase, both evidence chunks appear in the top-$w$ candidate pool, with the synopses showing they are about ``total revenues'' and ``property and equipment''. Therefore, the two are correctly selected. Then, the extraction phase successfully extracts necessary evidence from the raw chunks. Besides the raw data including the total revenue and both the beginning (2017) and ending (2018) PP\&E, the averaged PP\&E and the fixed asset turnover ratio are also calculated based on the information, which allows the generator LM to perform clear reasoning and generate the correct answer.

This case demonstrates SANE better selects the evidence with the help of the synopsis-based candidate selection, and has a better presentation of the information required for the query based on the blueprint-guided evidence extraction.

\section{Related Work}

\subsection{Retrieval-Augmented Generation}
RAG allows LM systems to generate with external information~\cite{RAG,rag-survey-24,rag-survey-26}. Existing studies have been mainly improving RAG from two directions. The first is retrieval quality, in which approaches like sparse~\cite{bm25}, dense~\cite{DPR}, hybrid~\cite{Hybrid-RAG}, late-interaction~\cite{ColBERT} retrievers, query rewriting~\cite{Query-Rewriting}, and document expansion~\cite{Doc2Query} are designed. The second direction builds index-time structures like trees~\cite{RAPTOR} or graphs~\cite{GraphRAG,HippoRAG,HippoRAG2} over the corpus. These methods mainly focus on the index-time, and the structures are usually constructed before knowing the query. Therefore, they can be expensive to build, and risk mismatching the needs of specific queries. SANE performs query-time augmentation like some recent work~\cite{StructRAG,RAS,SLIDERS}: after retrieving a broad candidate pool, we perform lightweight query-specific selection and evidence extraction before generation. This avoids heavy corpus-wide restructuring while providing the generator a more usable evidence interface.

\subsection{Post-Retrieval Selection}
It has been validated that LMs can be used to perform reranking very well for document search~\cite{ChatGPT-reranking,Listwise-reranking}. RankVicuna~\cite{RankVicuna} and RankZephyr~\cite{RankZephyr} train LMs specifically for reranking. All these methods use the full documents for reranking, which have limitations in both performance and cost (as shown in our ablation study). CoRank~\cite{CoRank} performs reranking based on LM-extracted features like category, sections, and keywords, which calls the LM for every document. How to perform efficient post-retrieval selection of candidates to mitigate the limitations of simple semantic retrievers is sparsely studied for RAG. Our method is fully training-free and employs extremely cheap features like existing pre-metadata (\textit{e.g.}, ID, date, page number), excerpts, and keywords to perform cost-friendly post-retrieval selection.

\subsection{Query-time Structuring}
There have been various approaches for query-time structuring. StructRAG~\cite{StructRAG} classifies queries and converts retrieved content into formats according to the query type. RAS~\cite{RAS} and SARG~\cite{SARG} construct query-specific knowledge graphs, which require trained models or a lot of query-time LM calls and do not support flexible schemas. SLIDERS~\cite{SLIDERS} extracts documents into relational databases based on query-specific schemas and employs a coding LM to reason on it, which requires numerous LM calls for extraction and strict database constraints. SchemaRAG~\cite{SchemaRAG} reconstructs all the documents based on query-specific schemas and performs retrieval and reasoning on that basis, which is very costly for each query. Most of these methods prioritize accuracy over efficiency and are typically heavy with a lot of pipeline steps and LM calls; some even require specific LM training. In realistic scenarios, efficiency is an important dimension to ensure a good user experience and system responsiveness. Our method for query-time structuring is training-free and does not need a lot of extra LM calls for each query, which balances performance and efficiency.

Overall, existing work leaves room for a lightweight query-time plugin that helps both candidate selection and evidence structuring, which is precisely the gap addressed by SANE.

\section{Discussion and Conclusion}
\subsection{Limitations}
Although SANE exhibits outstanding results with modest overhead, we acknowledge that our work has certain limitations. First, due to limited resource, we only perform experiments using two families of LMs as the backbone. We also cannot perform multiple algorithm runs due to the prohibitive cost of many methods. Second, the blueprints for the evidence extraction only cover three simple frequent scenarios, which may deserve further refinement.

\subsection{Future Work}
There are a few future directions worth exploration. First, further improve the blueprints by adding more high-quality schemas or designing lightweight blueprint selection mechanisms, so that the blueprints can be more flexible to adapt to different scenarios and domains. Second, integrate our method into practical agent systems (\textit{e.g.}, coding agent, computer-use agent) as a plugin of the default RAG tool to see its performance in agentic scenarios.

\subsection{Conclusion}
We have presented SANE, a simple yet effective plugin for RAG, which is plug-and-play, lightweight, and training-free. It performs synopsis-based candidate selection over a wide candidate pool, and then conducts blueprint-guided evidence extraction for the selected candidates for the query. Empirical results confirm that SANE brings solid improvements and outperforms all the compared methods while introducing only modest extra overhead, which can be significantly more efficient than many advanced methods. These results demonstrate that SANE brings a lightweight alternative to heavier methods and serves as a simple yet effective plugin for RAG. We suggest that an advanced RAG framework need not be overly complex, and hope our work could inspire future research on cost-efficient RAG approaches.


\bibliography{aaai2027}


\appendix

\section{More Implementation Details}

\subsection{Hyperparameters}
For all methods, we use their official default hyperparameters. If a method uses a retrieval budget such as top-$k$, we set it to the same $k$ as SANE.

For RAPTOR, its default hyperparameter could not run successfully on FinanceBench and LongMemEval. We therefore increased the maximum tokens per leaf node from 100 to 2,048, the maximum tokens per cluster from 3,500 to 32,768, and reduced the dimensionality from 10 to 5 to make it runnable on those benchmarks.

\subsection{Details of Blueprints}

\begin{table*}[htbp]
\small
  \centering
    \begin{tabularx}{\textwidth}{>{\hsize=.25\hsize\linewidth=\hsize}X
  >{\hsize=.5\hsize\linewidth=\hsize}X>{\hsize=1.2\hsize\linewidth=\hsize}X}
    \toprule
    \textbf{Column} & \textbf{Definition} & \textbf{Rules} \\
    \midrule
    fact  & A single atomic proposition expressed in one sentence & Split compound statements with "and"/"but"/"or" into separate rows. Use the entity or topic as the subject. \\
    \midrule
    evidence quote & Verbatim text from source & Exact quote. Use ... for elision. Never paraphrase. \\
    \midrule
    source & Which document/session & Consistent label across rows. \\
    \midrule
    source date & When the source was created & ISO 8601 when resolvable; raw expression otherwise. NULL if unknown. \\
    \bottomrule
    \end{tabularx}%
  \caption{The blueprint for ``ATOMIC''.}
  \label{tab:atomic}%
\end{table*}%

\begin{table*}[htbp]
\small
  \centering
    \begin{tabularx}{\textwidth}{>{\hsize=.25\hsize\linewidth=\hsize}X
  >{\hsize=.5\hsize\linewidth=\hsize}X>{\hsize=1.2\hsize\linewidth=\hsize}X}
    \toprule
    \textbf{Column} & \textbf{Definition} & \textbf{Rules} \\
    \midrule
    entity & Which thing is being described & Same canonical name in every row for the same entity. \\
    \midrule
    dimension & What attribute is being compared & Same dimension label for all entities on the same comparison axis. Rows sharing a dimension form a comparison group. \\
    \midrule
    value & The attribute value & Numeric when the source provides a number. Source text otherwise. NULL when this entity has no information for this dimension — asymmetry is information. \\
    \midrule
    unit  & Unit of measurement & Standardized (million, km, percent, USD). Empty for categorical or unquantified values. \\
    \midrule
    evidence quote & Verbatim text & Exact quote. \\
    \midrule
    source & Which document/session & — \\
    \bottomrule
    \end{tabularx}%
  \caption{The blueprint for ``COMPARATIVE''.}
  \label{tab:comparative}%
\end{table*}%

\begin{table*}[htbp]
\small
  \centering
    \begin{tabularx}{\textwidth}{>{\hsize=.25\hsize\linewidth=\hsize}X
  >{\hsize=.5\hsize\linewidth=\hsize}X>{\hsize=1.2\hsize\linewidth=\hsize}X}
    \toprule
    \textbf{Column} & \textbf{Definition} & \textbf{Rules} \\
    \midrule
    event & What happened & Brief description (5-15 words). Include the key entity or action. \\
    \midrule
    position & Ordinal position in the sequence & 1-based integer. Ties (same position number) are allowed and expected when order within a group is ambiguous. \\
    \midrule
    resolved date & Best-guess absolute date & ISO 8601 (YYYY-MM-DD). NULL when unresolvable — this is common and acceptable. \\
    \midrule
    date precision & How precisely the date is known & day (exact), month (month known, day unknown), year (year only), relative (only order known, not absolute date). \\
    \midrule
    evidence quote & Verbatim text & Exact quote. \\
    \midrule
    source & Which document/session & — \\
    \bottomrule
    \end{tabularx}%
  \caption{The blueprint for ``TEMPORAL''.}
  \label{tab:temporal}%
\end{table*}%

The detailed definitions of the three blueprints used by SANE are presented in Tables \ref{tab:atomic}, \ref{tab:comparative}, and \ref{tab:temporal}, respectively. These definitions are included in the extraction LM's context.

\subsection{Prompt Templates for the LMs}

\begin{figure*}
\begin{promptbox}{Vanilla RAG}
I will give you several history records.
Please answer the question based on the relevant records.

History Records:
\{history\}

Current Date: \{question\_date\}
Question: \{question\}
Answer:
\end{promptbox}
\caption{The prompt template for the vanilla RAG.}
\label{fig:prompt-rag}
\end{figure*}

\begin{figure*}
\begin{promptbox}{Synopsis-Based Candidate Selection}
\emph{System:}\\
You are a document selector for a retrieval system. Given a question and synopses of candidate documents, pick the ones most likely to contain information relevant to answering the question.

Each line shows a quoted excerpt and keyword tags.

Output ONLY a JSON list of up to \{max\_select\} line numbers (1-based).\\
Example: [2, 5, 7]\\
Example (no relevant docs): []

\medskip
\emph{User:}\\
\#\# Question\\
\{query\}

\#\# Candidate Sessions (sorted by embedding similarity)\\
\{synopses\}

Output a JSON list of up to \{max\_select\} session numbers that are most relevant to the question.
\end{promptbox}
\caption{The prompt template for the synopsis-based candidate selection of SANE.}
\label{fig:prompt-selection}
\end{figure*}

\begin{figure*}
\begin{promptbox}{Blueprint-Guided Evidence Extraction}
\emph{System:}\\
You are a precise evidence extractor. Given a question and retrieved content, extract ONLY the information needed to answer the question. The schemas below are design inspiration — useful starting points, not rigid templates.

\#\# Schema Lookup Table

| Schema | Information Structure | Use When | Columns |\\
|---|---|---|---|\\
| ATOMIC | Independent facts — no row relates to any other |
  What/Who/Where/Define/Find/Yes-no questions |
  fact, evidence\_quote, source, source\_date |\\
| COMPARATIVE | Aligned dimensions — entities compared on shared
  attributes | Compare/Difference/Versus/Which is larger-better-faster/Rank |
  entity, dimension, value, unit, evidence\_quote, source |\\
| TEMPORAL | Sequenced events — rows ordered by occurrence time |
  When/First/Last/Most recent/Timeline/Before-After/Sequence |
  event, position, resolved\_date, date\_precision, evidence\_quote, source |

\#\# Column Definitions

\#\#\# ATOMIC\\
\{atomic\_blueprint\}

\#\#\# COMPARATIVE\\
\{comparative\_blueprint\}

\#\#\# TEMPORAL\\
\{temporal\_blueprint\}

\#\# How to Use the Schemas

- Schemas are design aids, not constraints. The goal is a clear evidence table — not schema compliance.\\
- Use a schema directly when it fits the question well.\\
- Adapt columns freely — rename domain-specific ones, drop irrelevant ones, borrow from multiple schemas.\\
- When the evidence is sparse or the question doesn't map cleanly to any schema, a simple fact table is perfectly fine.\\
- When multiple perspectives apply (e.g., comparison + temporal order), you may blend columns or extract multiple complementary tables.

\#\# Instructions

1. Read the question carefully. Identify what information structure best fits: independent facts, aligned comparison, or sequenced events.

2. Consult the schemas above for inspiration. Use them as flexible starting points. Let the retrieved content shape the schema as much as the blueprints do.

3. Scan the retrieved content. Find every piece of relevant information. IGNORE everything irrelevant.

4. Design a schema that fits the evidence you found. Be exhaustive — the answerer will filter and reason.

5. Every row MUST include an exact evidence\_quote. Never paraphrase.

6. Output ONLY the markdown table. No preamble, no commentary, no answer.

\medskip
\emph{User:}\\
\#\# Question\\
\{query\}

\#\# Retrieved History\\
\{formatted\_history\}

Extract the minimum evidence needed to answer this question as a markdown table.
\end{promptbox}
\caption{The prompt template for the blueprint-guided evidence extraction of SANE.}
\label{fig:prompt-extraction}
\end{figure*}

\begin{figure*}
\begin{promptbox}{Answer Generation}
I will give you extracted evidence from history records.\\
Please answer the question based on this evidence.

Extracted Evidence:\\
\{evidence\}

Current Date: \{question\_date\}\\
Question: \{question\}\\
Answer:
\end{promptbox}
\caption{The prompt template for the answer generation of SANE.}
\label{fig:prompt-sane-generation}
\end{figure*}

\begin{figure*}
\begin{promptbox}{Full-Text Candidate Selection}
You are a document selector for a retrieval system. Given a question and full-text records, pick the ones most likely to contain information relevant to answering the question.

Each record is shown as a \#\# Record N section with its full content.

Output ONLY a JSON list of up to \{max\_select\} record numbers (1-based).\\
Example: [2, 5, 7]\\
Example (no relevant docs): []
\end{promptbox}
\caption{The prompt template for the full-text candidate selection variant (\textit{i.e.}, ``w/o synopsis'').}
\label{fig:prompt-full-text}
\end{figure*}

\begin{figure*}
\begin{promptbox}{Vanilla Information Extraction}
\emph{System:}\\
You are a precise evidence extractor. Given a question and retrieved content, extract ONLY the information needed to answer the question.

Output ONLY the extracted information.

\medskip
\emph{User:}\\
\#\# Question\\
\{query\}

\#\# Retrieved History\\
\{formatted\_history\}

Extract the minimum evidence needed to answer this question as a markdown table.
\end{promptbox}
\caption{The prompt template for the vanilla information extraction variant (\textit{i.e.}, ``w/o blueprint'').}
\label{fig:prompt-no-blueprint}
\end{figure*}

The generator's prompt template for Embedding (vanilla RAG) is shown in Figure~\ref{fig:prompt-rag}.

For SANE, the prompt templates for synopsis-based candidate selection, blueprint-guided evidence extraction, and answer generation are presented in Figures \ref{fig:prompt-selection}, \ref{fig:prompt-extraction}, and \ref{fig:prompt-sane-generation}, respectively. The template for the ``w/o synopsis'' ablation is in Figure~\ref{fig:prompt-full-text}, and that for the ``w/o blueprint'' ablation is in Figure~\ref{fig:prompt-no-blueprint}.

\subsection{Other Details}
For StructRAG, we train Qwen3.5-9B with LoRA~\cite{LoRA} as the router for both backbone LMs in the main experiment. Due to compatibility issues, we use standard supervised fine-tuning (SFT) instead of the official direct preference optimization (DPO)~\cite{DPO}. More specifically, DPO cannot override the LM's reasoning tokens, so it fails to train the router as intended.

\section{Extended Experimental Results}

\subsection{Significance of Improvement}

\begin{table*}[htbp]
\centering
\begin{tabular}{lrrrr}
\toprule
\textbf{Method} & \textbf{HotpotQA} & \textbf{MuSiQue} &
\textbf{FinanceBench} & \textbf{LongMemEval} \\
\midrule
\multicolumn{5}{c}{\textit{DeepSeek-V4-Flash}} \\
\midrule
Embedding & +9.78$^{***}$ & +12.17$^{***}$ & +16.66$^{***}$ & +14.20$^{***}$ \\
RAPTOR & +12.88$^{***}$ & +15.19$^{***}$ & +30.00$^{***}$ & +27.60$^{***}$ \\
HippoRAG 2 & +5.52$^{***}$ & +7.49$^{***}$ & +16.66$^{***}$ & +7.80$^{**}$\phantom{$^{*}$} \\
StructRAG & +14.63$^{***}$ & +18.83$^{***}$ & +14.66$^{***}$ & +18.80$^{***}$ \\
RAS$_\text{lite}$ & +19.62$^{***}$ & +13.53$^{***}$ & +42.66$^{***}$ & +42.20$^{***}$ \\
\midrule
\multicolumn{5}{c}{\textit{Qwen3.5-9B}} \\
\midrule
Embedding & +13.27$^{***}$ & +18.33$^{***}$ & +14.00$^{**}$\phantom{$^{*}$} & +16.40$^{***}$ \\
RAPTOR & +18.43$^{***}$ & +21.18$^{***}$ & +36.66$^{***}$ & +31.00$^{***}$ \\
HippoRAG 2 & +7.44$^{***}$ & +9.68$^{***}$ & +14.66$^{**}$\phantom{$^{*}$} & +13.20$^{***}$ \\
StructRAG & +25.07$^{***}$ & +32.11$^{***}$ & +28.00$^{***}$ & +37.60$^{***}$ \\
RAS$_\text{lite}$ & +40.18$^{***}$ & +39.47$^{***}$ & +47.33$^{***}$ & +47.20$^{***}$ \\
\bottomrule
\end{tabular}
\caption{Improvement of SANE over each baseline in percentage points. Significance is assessed using paired McNemar tests. $^{*}p<0.05$, $^{**}p<0.01$, $^{***}p<0.001$.}
\label{tab:significance}
\end{table*}

In the main experiments, the significance of SANE's improvement over baseline methods is judged using McNemar's test (Table~\ref{tab:significance}). All reported $p$-values are below 0.001, except SANE versus HippoRAG 2 on LongMemEval with DeepSeek-V4-Flash (p=0.0037) and SANE versus Embedding (p=0.0014) and HippoRAG 2 (p=0.0012) on FinanceBench with Qwen3.5-9B.

\subsection{Effect of Retrieval Hyperparameters}

\begin{table}[htbp]
  \centering
    \begin{tabular}{lrrrr}
    \toprule
    \textbf{Setting} & \textbf{MQ} & \textbf{FB} & \textbf{LME} & \textbf{Avg.} \\
    \midrule
    RAG (k=3) & 21.6  & 48.7  & 36.2  & 35.5 \\
    SANE (k=3, w=20) & \textbf{40.8} & \textbf{68.7} & \textbf{54.8} & \textbf{54.7} \\
    \midrule
    RAG (k=10) & 33.0  & 61.3  & 47.6  & 47.3 \\
    SANE (k=10, w=20) & \textbf{46.6} & \textbf{68.7} & \textbf{54.8} & \textbf{56.7} \\
    \midrule
    RAG (k=5) & 26.5  & 57.3  & 40.0  & 41.3 \\
    SANE (k=5, w=10) & 38.8  & 60.0  & 49.4  & 49.4 \\
    SANE (k=5, w=20) & \textbf{44.9} & \textbf{71.3} & \textbf{56.4} & \textbf{57.5} \\
    SANE (k=5, w=30) & 44.7  & 70.7  & 49.4  & 54.9 \\
    \bottomrule
    \end{tabular}%
  \caption{Effect of retrieval hyperparameters. Evaluated on MuSiQue (MQ), FinanceBench (FB), and LongMemEval (LME), with Qwen3.5-9B as the backbone LM.}
  \label{tab:hyperparameters}%
\end{table}%

We analyze the effect of retrieval hyperparameters ($w$ and $k$) in Table~\ref{tab:hyperparameters}. SANE consistently improves performance across all combinations of $w$ and $k$.

Regarding $w$, we keep $k$ fixed at 5 and evaluate with three $w$ values: 10, 20 (default), and 30. The results suggest that a smaller $w$ leads to a substantial performance drop, which is reasonable because recall@$w$ is naturally lower when $w$ is small. Meanwhile, a larger $w$ does not necessarily help, possibly because too many candidates overload the selection LM's context.

Regarding $k$, SANE degrades gracefully with a smaller $k$, while a larger $k$ further benefits MuSiQue but slightly hurts FinanceBench and LongMemEval, possibly because the extraction LM's context becomes overloaded.

Overall, the current $w$ and $k$ values are near optimal, and SANE yields consistent gains over vanilla RAG across all evaluated settings.

\subsection{Extraction Quality Breakdown}

\begin{table}[htbp]
  \centering
    \begin{tabular}{lrr}
    \toprule
    \textbf{Classification} & \multicolumn{1}{l}{\textbf{Count}} & \multicolumn{1}{l}{\textbf{Percentage}} \\
    \midrule
    Faithful & 615   & 81.1\% \\
    \midrule
    Partial & 26    & 3.4\% \\
    Error & 12    & 1.6\% \\
    Hallucinated & 105   & 13.9\% \\
    \bottomrule
    \end{tabular}%
  \caption{Distribution of extracted rows.}
  \label{tab:faithfulness}%
\end{table}%

We quantitatively evaluate the quality of blueprint-guided evidence extraction on FinanceBench using Qwen3.5-9B as the backbone LM. We prompt an advanced LM, DeepSeek-V4-Pro \cite{DeepSeek-V4}, to classify each row in the extracted information into four categories: (1) Faithful: directly supported by the retrieved content; (2) Partial: only partially supported; (3) Error: derived from incorrect reasoning; (4) Hallucinated: contradicted or unsupported by the retrieved content (typically fabricated).

As presented in Table~\ref{tab:faithfulness}, 81.1\% of the extracted rows are faithful, and only 5.0\% rows are partial or error, showing the blueprint-guided extraction is largely reliable. Nevertheless, there are 13.9\% rows hallucinated. We acknowledge this is a limitation of the single-pass extraction, and leave more faithful single-pass extraction strategies for future work.

\subsection{Distribution of Extracted Schemas}

\begin{table}[htbp]
  \centering
    \begin{tabular}{lrrr}
    \toprule
    \textbf{Schema} & \multicolumn{1}{l}{\textbf{Count}} & \multicolumn{1}{l}{\textbf{Percentage}} & \multicolumn{1}{l}{\textbf{\#Rows}} \\
    \midrule
    ATOMIC & 117   & 78.0\% & 5.1 \\
    COMPARATIVE & 22    & 14.7\% & 8.5 \\
    TEMPORAL & 2     & 1.3\% & 7.0 \\
    OTHER & 9     & 6.0\% & 4.4 \\
    \bottomrule
    \end{tabular}%
  \caption{Distribution of extracted schemas. ``\#Rows'' shows the averaged number of rows per extracted table.}
  \label{tab:schema-distribution}%
\end{table}%

Table~\ref{tab:schema-distribution} shows the distribution of the extracted schemas on FinanceBench with Qwen3.5-9B as the backbone LM. ``ATOMIC'' dominates most cases, whereas ``TEMPORAL'' is rare. Notably, the extraction LM is not constrained by the blueprints and can design an appropriate schema on its own when necessary, as shown by the 6\% ``OTHER'' schemas.

\subsection{Full Results across Backbone LMs}

\begin{table*}[htbp]
  \centering
    \begin{tabular}{llrrrrr}
    \toprule
    \textbf{Backbone LM} & \textbf{Method} & \textbf{HQ} & \textbf{MQ} & \textbf{FB} & \textbf{LME} & \textbf{Avg.} \\
    \midrule
    \multirow{2}[0]{*}{Qwen3.5-4B} & Embedding & 43.04 & 21.85 & 57.33 & 39.2  & 40.36 \\
          & SANE  & \textbf{56.16} & \textbf{37.77} & \textbf{71.33} & \textbf{50.4} & \textbf{53.92} \\
    \midrule
    \multirow{2}[0]{*}{Qwen3.5-9B} & Embedding & 48.40 & 26.52 & 57.33 & 40.0  & 43.06 \\
          & SANE  & \textbf{61.67} & \textbf{44.85} & \textbf{71.33} & \textbf{56.4} & \textbf{58.56} \\
    \midrule
    \multirow{2}[0]{*}{Qwen3.6-35B-A3B-FP8} & Embedding & 51.68 & 32.31 & 58.67 & 45.0  & 46.92 \\
          & SANE  & \textbf{63.55} & \textbf{50.35} & \textbf{70.67} & \textbf{59.0} & \textbf{60.89} \\
    \midrule
    \multirow{2}[0]{*}{DeepSeek-V4-Flash} & Embedding & 50.18 & 28.09 & 58.67 & 43.2  & 45.04 \\
          & SANE  & \textbf{59.96} & \textbf{40.26} & \textbf{75.33} & \textbf{57.4} & \textbf{58.24} \\
    \midrule
    \multirow{2}[0]{*}{DeepSeek-V4-Pro} & Embedding & 53.99 & 34.38 & 60.00 & 44.6  & 48.24 \\
          & SANE  & \textbf{63.15} & \textbf{48.94} & \textbf{69.33} & \textbf{60.8} & \textbf{60.56} \\
    \bottomrule
    \end{tabular}%
  \caption{Full results comparing Embedding and SANE on HotpotQA (HQ), MuSiQue (MQ), FinanceBench (FB), and LongMemEval (LME), with multiple different backbone LMs.}
  \label{tab:all-models}%
\end{table*}%

Table~\ref{tab:all-models} shows the full results comparing Embedding and SANE with multiple different backbone LMs. SANE consistently brings significant gains over the vanilla RAG across all five backbone LMs and all four benchmarks. This shows that SANE benefits RAG whenever the LM is small (4 billion parameters for Qwen3.5-4B) or large (1.6 trillion total parameters for DeepSeek-V4-Pro) and the architecture is dense (Qwen3.5 series) or sparse (mixture-of-experts for Qwen3.6-35B-A3B-FP8 and the DeepSeek series).

\section{Computing Infrastructure}
The experiments were mainly run on a machine with an Apple M4 10-core processor and 16 GB of memory running macOS Tahoe 26.5.2.

For the DeepSeek-V4-Flash and DeepSeek-V4-Pro \cite{DeepSeek-V4} LMs, we use the official API provided by DeepSeek.

For Qwen LMs including Qwen3.5-4B, Qwen3.5-9B, and Qwen3.6-35B-A3B \cite{Qwen3}, we deploy them on 2$\times$ NVIDIA RTX PRO 6000 Blackwell Server Edition (96 GB) GPUs with vLLM 0.23.0 on a machine with Ubuntu 22.04.

For the embedding model, \textit{i.e.}, Qwen3-Embedding-0.6B \cite{qwen-embedding}, we deploy it on a single NVIDIA GeForce RTX 4090 (48 GB) GPU with vLLM v0.20.0 on a machine with Ubuntu 22.04.

All GPU resources were leased through AutoDL.

\section{Use of AI Systems}
The algorithmic design and core methodology of this work were derived through manual research and human reasoning. We use generative AI assistants solely for wording, editing, formatting, and coding purposes. They do not impact the scientific rigor or originality of the research.

\end{document}